\newif\ifTechReport
\TechReporttrue
\ifTechReport
\documentclass[fleqn,leqno,12pt,twoside]{nictatechreport}
\else
\documentclass[11pt,fleqn]{article}
\fi

\usepackage{amsmath}
\usepackage{amssymb}

\usepackage[utf8]{inputenc}
\usepackage[T1]{fontenc}

\usepackage{mathptmx}
\usepackage{libertine}
\let\models\vDash

\usepackage{microtype}
\usepackage{enumitem}
\usepackage{xspace}

\usepackage{etoolbox}
\newtoggle{TechReport}
\ifTechReport
\toggletrue{TechReport}
\else
\togglefalse{TechReport}
\fi

\usepackage[backend=bibtex,
            maxbibnames=99,
            isbn=false,
            ]{biblatex}            
\AtEveryBibitem{
  \iffieldundef{pages}{}{\clearfield{doi}\clearfield{url}}
  \clearfield{venue}
  \clearfield{eventyear}
  }
\renewbibmacro{in:}{%
  \ifentrytype{article}{}{\printtext{\bibstring{in}\intitlepunct}}}

\usepackage[multiple]{footmisc}
\usepackage{hyperref}

\newcommand{\set}[1]{\{#1\}}

\setlist{noitemsep,topsep={.25\abovedisplayskip}}

\title{Thou Shalt is not You Will}
\author{Guido Governatori}

\iftoggle{TechReport}{
\affiliation{Software Systems Research Group, NICTA, Queensland Research Lab, Brisbane, Australia \\
  Queensland University of Technology, Brisbane, Australia\\
  email: \texttt{guido.governatori@nicta.com.au}}
\reportnumber{8026}}
{}

\newcommand{\G}{\ensuremath{\mathsf{G}}\xspace}
\newcommand{\F}{\ensuremath{\mathsf{F}}\xspace}
\newcommand{\X}{\ensuremath{\mathsf{X}}\xspace}
\newcommand{\PER}{\ensuremath{\mathsf{P}}\xspace}

\newcommand{\OBL}{\ensuremath{\mathsf{O}}\xspace}
\newcommand{\U}{\ensuremath{\mathbin{\mathsf{U}}}}
\newcommand{\W}{\ensuremath{\mathbin{\mathsf{W}}}}
\newcommand{\forb}{\ensuremath{\mathop{\mathsf{Forbidden}}}}
\newcommand{\obl}{\ensuremath{\mathop{\mathsf{Obligatory}}}}
\newcommand{\per}{\ensuremath{\mathop{\mathsf{Permitted}}}}

\newcommand{\Nat}{\mathbb{N}}

\begin{document}

\iftoggle{TechReport}{}{\maketitle}

\begin{abstract}
 In this paper we discuss some reasons why temporal logic might not
 be suitable to model real life norms. To show this, we present a 
 novel deontic logic contrary-to-duty/derived permission paradox 
 based on the interaction of obligations, permissions and contrary-to-duty
 obligations. The paradox is inspired by real life norms. 
\end{abstract}

\iftoggle{TechReport}{

\begin{pubhistory}\ \\
2014-04-07 (Version 2, Revision 6)\\
2014-05-13 (Version 3, Revision 16)\\
2014-09-28 (Version 4, Revision 22)\\
2015-01-19 (Version 5, Revision 30)
\end{pubhistory}
\frontmatter
\mainmatter}{}

\section{Introduction}
\label{sec:intro}

The aim of this note is to discuss the reasons why temporal logic,
specifically Linear Temporal Logic \cite{Pnueli:1977} might not be suitable to check whether
the specifications of a system comply with a set of normative requirements.

The debate whether it is possible to use temporal logic for the representation 
of norms is not a novel one (see for example \cite{thomason}), and while the
argument had settled for a while, the past decade saw a resurgence of the 
topic with many works in the fields of normative multi-agents and business 
process compliance advocating temporal logic as the formalism to express 
normative constraints on agent behaviours and process executions.  One 
of the reasons behind this could be the success of model checking for 
temporal logic in verifying large scale industry applications\footnote{The 
fathers of model checking for temporal logic, i.e., Edmund Clarke,  
E. Allen Emerson and Joseph Sifakis, were the recipient of the Turing award 
in 2007 for their role in developing Model-Checking into a highly effective 
verification technology that is widely adopted in the hardware and software 
industries.}. 

The problem in normative multi-agents systems and business process compliance is 
to determine whether the actions an agent is going to perform (encoded as a plan,
corresponding to a sequence of actions) or the tasks to be executed by a business 
process conform with a set of normative constraints regulating their possible
(legal) behaviours. In both cases we have sequences of actions/tasks leading to 
sequences of states and constrains over what states and sequences of states are 
deemed legal according to a set of normative constrains. From a formal point 
of view both the behaviours and the constraints are represented by temporal 
logic formulas and an agent or process are compliant if the set of formulas 
is consistent.  Temporal logic is definitely capable to model the sequences 
of states corresponding the behaviours of agents and processes, but the issue
whether it is able to represents normative constraints (i.e., obligations 
and prohibition) in a conceptually sound way has been neglected. We believe 
that this is crucial issue to be addressed before these techniques can be 
proposed for practical real life cases.  Without a positive answer the work 
based on temporal logic for the representation of norms remains a futile formal 
exercise. 
 
The short discussion above boils down to the following question:
\begin{quote}
Are normative constrains (i.e., obligations and prohibitions) regulating the 
behaviours of their subjects different from other types of constraints?
\end{quote}
In case of a negative answer we have to identify what are the differences, and 
how to model them in temporal logic. Furthermore, we have to identify what 
are the issues with the resulting modelling. 

Obligations and prohibitions are constraints that limit the scope of 
actions of the bearer subject to them. However, there is a very important
difference between obligations and prohibitions and other types of constraints:
violations do not result in inconsistencies. This means that they can be 
violated without breaking the systems in which they appear. Accordingly, a
better understanding of obligations and prohibitions is that they define what
is legal (in a particular system) and what is illegal. Based on this reading 
a violation simply indicates that we ended up in an illegal situation or state.
A further aspect we have to consider, and that has been by large neglected by
investigations on how to formalise and reason with deontic concepts, is that
violations can be compensated for, and a situation where there is a violation
but there is a compensation for the violation is still deemed legal (even if,
from a legal point of view, less ideal than the situation where the violation
does not occur).  


The paper is organised as follows: in the next section we introduce a legal
scenario (a fragment of an hypothetical privacy act) illustrating some of
the aspects differentiating norms form other types of constraints, and we 
shortly discuss what the outcomes of cases related to this scenario should be.
Then in Section~\ref{sec:logic:background} we briefly recall the basics of 
Linear Temporal Logic (LTL). In Section~\ref{sec:logic:formalisation} we discuss 
how to formalise the scenario in LTL. We point out various shortcomings for 
the representation of norms in LTL, and we show that LTL captures only some
of the aspects of the scenario (suggesting that it is not able to model 
real life norms), or it leads to paradoxical results.\footnote{Following \citeauthor{aqvist}’s 
\cite{aqvist} presentation, a paradox arises in a deontic logic $\Delta$ either when 
there is a formula $\phi$ derivable in $\Delta$ but for which the translation 
does not seem derivable within the natural normative language, or there is a 
formula $\phi$ which is not derivable in $\Delta$ but for which the translation 
seems derivable within our natural normative language.}

\section{Legal Motivation}
\label{sec:scenario}

Suppose that a Privacy Act contains the following norms:\footnote{The 
Privacy Act presented here, though realistic, is a fictional one. However, 
(i) it is based on the novel Australian Privacy Principles (APP), Privacy Amendment (Enhancing Privacy Protection) Act 2012, and (ii) sections 
with the same logical structure as the clauses of this fictional act are 
present in the APP Act.}
\begin{enumerate}[label={Section \arabic{*}.}, leftmargin=*]
\item The collection of personal information is forbidden, unless acting on
  a court order authorising it.
\item The destruction of illegally collected personal information before
  accessing it is a defence against the illegal collection of the personal
  information.
\item The collection of medical information is forbidden, unless the entity
  collecting the medical information is permitted to collect personal
  information.
\end{enumerate}
In addition the Act specifies what personal information and medical
information are, and they turn out to be disjoint.

Suppose an entity, subject to the Act, collects some personal information
without being permitted to do so; at the same time they collect medical
information. The entity recognises that they illegally collected personal
information (i.e., they collected the information without being authorised to
do so by a Court Order) and decides to remediate the illegal collection by
destroying the information before accessing it. Is the entity
compliant with the Privacy Act above? Given that the personal information was
destroyed the entity was excused from the violation of the first section (illegal
collection of personal information). However, even if the entity was excused
from the illegal collection, they were never entitled (i.e., permitted) to
collect personal information\footnote{If they were permitted to collect
personal information, then the collection would have not been illegal, and 
they did not have to destroy it.}, consequently they were not permitted to 
collect medical information; thus the prohibition of collecting medical 
information was in force. Accordingly, the collection of medical information 
violates the norm forbidding such an activity.

Let us examine the structure of the act:

Section 1 establishes two conditions: 
\begin{enumerate}[label={\roman{*}.}]
\item Typically the collection of personal  information is forbidden; and
\item The collection of personal information is permitted, if there is a
  court order authorizing the collection of personal information.
\end{enumerate}
Section 2 can be paraphrased as follows:
\begin{enumerate}[label={\roman{*}.},resume]
\item The destruction of personal information collected illegally before accessing it excuses the illegal collection.  
\end{enumerate}
Similarly to Section 2, Section 3 states two conditions:
\begin{enumerate}[label={\roman{*}.},resume]
\item Typically the collection of medical information is forbidden; and
\item The collection of medical information is permitted provided that the
  collection of personal information is permitted.
\end{enumerate}
Based on the above discussion, if we abstract from the actual content of the
norms, the structure of the act can be represented by the following set of
norms (extended form):
\begin{enumerate}[label={E\arabic{*}.}]
\item $A$ is forbidden. 
\item $A$ is permitted given $C$ (alternatively: if $C$, then $A$ is permitted).
\item The violation of $A$ is compensated by $B$
\item $D$ is forbidden.
\item If $A$ is permitted, so is $D$.
\end{enumerate}
To compensate a violation we have to have a violation the compensation
compensates. Moreover, to have a violation we have to have an obligation or
prohibition, the violation violates. Accordingly, it makes sense to combine E1
and E3 in a single norm, obtaining thus the following set of norms (condensed
form):
\begin{enumerate}[label={C\arabic{*}.}]
\item $A$ is forbidden; its violation is compensated by $B$.
\item $A$ is permitted given $C$ (alternatively: if $C$, then $A$ is permitted).
\item $D$ is forbidden.
\item If $A$ is permitted, so is $D$.
\end{enumerate}
Based on the discussion so far the logical structure of the act is (logical
form):

\begin{enumerate}[label={L\arabic{*}.}]
\item $\forb A$;  if $\forb A$ and $A$, then $\obl B$.
\item if $C$, then $\per A$.
\item $\forb D$.
\item If $\per A$, then $\per D$.
\end{enumerate}
Notice the way we modelled the violation of the prohibition of $A$ in L1,
namely as the conjunction of $A$ and the prohibition of
$A$.\footnote{Similarly, the violation of the obligation of $A$ is the
conjunction of obligation $A$ and the negation of the content of the
obligation, that is, $\neg A$.} Then we model that $B$ is the compensation of
the violation of $A$ as an implication from the violation of $A$ to the
obligation of $B$.

Let us consider what are the situations compliant with the above set of norms. 
Clearly, if $C$ does not hold, then we have that the prohibition of $A$ 
and prohibition of $D$ are in force. Therefore, a situation where $\neg A$, $\neg C$, and $\neg D$
hold is fully compliant (irrespective whether $B$ holds or not).
If $C$ holds, then the permission of $A$ derogates the prohibition of $A$, thus
situations with either $A$ holds or $\neg A$ holds are compliant with the first two
norms; in addition, the permission of $A$ allows us to derogate the
prohibition of $D$. Accordingly, situations with either $D$ or $\neg D$ comply
with the third norm. Let us go back to scenarios where $C$ does not hold, and
let us suppose that we have $A$. This means that the prohibition of $A$ has
been violated; nevertheless the set of norms allows us to recover from such
a violation by $B$. However, as we just remarked above
to have a violation we have to have either an
obligation or a prohibition that has been violated: in this case the
prohibition of $A$. Given that the prohibition of $A$ and the permission of $A$ are
mutually incompatible, we must have, to maintain a consistent situation, that
$A$ is not permitted. But if $A$ was not permitted $D$ is not permitted
either; actually, according to the third norm, $D$ is forbidden. To sum up, a
scenario where $\neg C$, $A$, $B$ and $\neg D$ hold is still compliant (even
if to a lesser degree given the compensated violation of the prohibition of
$A$). In any case, no situation where both $\neg C$ and $D$ hold is compliant.

Table~\ref{tab:compliance_status} summarises the compliant and not compliant
situations. We only report the minimal sets required to identify whether a
situation is compliant or not. For non-minimal sets the outcome is determined
by the union of the status for the minimal subsets.

\begin{table}[htb]
\centering
\begin{tabular}{ll}
  \textbf{Minimal Set} & \textbf{Compliance Status}\\
  \hline 
  $C$ & compliant\\
  $\neg C$, $A$, $B$ 
    & weakly compliant: compensated violation of the prohibition of $A$\\
  $\neg C$, $A$, $\neg B$
    & not compliant: uncompensated violation of the prohibition of $A$\\
  $\neg C$, $D$ & not compliant: violation of prohibition of $D$\\
  $\neg C$, $\neg A$, $\neg D$ & compliant
\end{tabular}
\caption{Compliance Status for the Privacy Act}
\label{tab:compliance_status}
\end{table}

\section{Logic  Background}
\label{sec:logic:background}

Linear Temporal Logic \cite{Pnueli:1977} is equipped with three unary temporal operators: 
  \begin{itemize}
  \item $\X\phi$: next $\phi$ ($\phi$ holds at the next time);
  \item $\F\phi$: eventually $\phi$ ($\phi$ holds sometimes in the future); and
  \item $\G\phi$: globally $\phi$ ($\phi$ always holds in the future).
\end{itemize}
In addition we have the following binary operators: 
\begin{itemize}
  \item $\phi\U\psi$:  $\phi$ until $\psi$  ($\phi$ holds until $\psi$ holds);
  \item $\phi\W\psi$: $\phi$ weak until $\psi$ ($\phi$ holds until $\psi$
    holds and $\psi$ might not hold).
\end{itemize}
The operators above are related by the following equivalences establishing some
 interdefinability among them:
\begin{itemize}
  \item $\F\phi\equiv \top\U\phi$,
  \item $\G\phi\equiv\neg\F\neg\phi$,
  \item $\phi\W\psi \equiv (\phi\U\psi)\vee\G\phi$.
\end{itemize}
The semantics of LTL can be given in terms of transition systems. A \emph{transition system} $TS$ is a structure
\begin{equation}\label{eq:transition}
  TS=\langle S,R,v\rangle
\end{equation}
where
\begin{itemize}
  \item $S$ is a (non empty) set of states
  \item $R\subseteq S \times S$ such that 
  \(
    \forall s\in S\exists t\in S\colon (s,t)\in R
  \)
  \item $v$ is a valuation function $v\colon S\mapsto 2^\mathit{Prop}$ 
\end{itemize}
where $\mathit{Prop}$ is the set of atomic propositions. 

Formulas in LTL are evaluated against fullpaths (also called traces or runs). 
A \emph{fullpath} is a sequence of states in $S$ connected by the transition
relation $R$. Accordingly, $\sigma=s_0, s_1, s_2\dots$ is a fullpath if and 
only if $(s_i,s_{i+1})\in R$. Given a fullpath $\sigma$, $\sigma_i$  denotes
the subsequence of $\sigma$ starting from the $i$-th element, and $\sigma[i]$
denotes the $i$-th element of $\sigma$.

Equipped with the definitions above, the valuation conditions for the various 
temporal operators are:
\begin{itemize}
  \item $TS,\sigma\models p$ $(p\in\mathit{Prop})$ iff $p\in v(\sigma[0])$;
  \item $TS,\sigma\models \neg\phi$ iff $TS,\sigma\not\models\phi$;
  \item $TS,\sigma\models \phi\wedge\psi$ iff $TS,\sigma\models\phi$ and
    $TS,\sigma\models\psi$;
  \item $TS,\sigma\models\X\phi$ iff $TS,\sigma_{1}\models\phi$;
  \item $TS,\sigma\models\phi\U\psi$ iff $\exists k\colon k\geq0,\
    TS,\sigma_k\models\psi$ and $\forall j\colon 0\leq j<k$,
    $TS,\sigma_{j}\models\phi$;
  \item $TS,\sigma\models \G\phi$ iff $\forall k\geq 0,\ TS,\sigma_k\models\phi$;
  \item $TS,\sigma\models \F\phi$ iff $\exists k\geq 0,\ TS,\sigma_k\models\phi$.
\end{itemize}
A formula $\phi$ is true in a fullpath $\sigma$ iff it is true at the first
element of the fullpath. Next we define what it means for a formula  $\phi$ 
to be true in a state $s\in S$ ($TS,s\models\phi$).
\begin{equation}
  TS,s \models \phi \text{ iff } \forall \sigma\colon \sigma[0]=s,\
  TS,\sigma\models\phi.
\end{equation}

\section{Scenario Formalised}
\label{sec:logic:formalisation}
The first problem we have to address is how to model obligations and
permissions in Linear Temporal Logic. When one considers the temporal
lifecycle obligations, obligations can be classified as \emph{achievement} and
\emph{maintenance} obligations \cite{it:bpc:anf}. After an obligation enters into force, the
obligation remains in force for an interval of time. A maintenance obligation
is an obligation whose content must hold for every instant in the interval in
which the obligation is in force. On the other hand, for an achievement
obligation, the content of the obligation has to hold at least once in the
interval of validity of the obligation.  Accordingly, a possible solution is to
use $\G$ to model maintenance obligations\footnote{We can use $\U$ instead 
of $\G$ to capture that an obligation is in force in an interval.} and $\F$ for 
achievement obligations. A drawback of this proposal is that $\G$ and $\F$ 
are the dual of each other, i.e., $\G\alpha\equiv\neg\F\neg\alpha$. In Deontic
Logic permission is typically defined as the lack of the obligation to the 
contrary and the deontic operators $\OBL$ and $\PER$ to model obligations and 
permissions are defined to be the dual of each other, namely 
$\OBL\alpha\equiv\neg\PER\neg\alpha$. In addition, most deontic logics
assume the following axiom (Axiom D)\footnote{In terms of Kripke possible 
world semantics Axiom D is characterised by \emph{seriality}, i.e., 
$\forall x\exists y(xRy)$, and this is the property imposed on the transition 
relation $R$ over the set of states $S$ in a transition system for LTL.}
\begin{equation}
  \OBL\alpha\rightarrow\PER\alpha
\end{equation}
to ensure consistency of sets of norms. The axiom is equivalent to 
$\OBL\alpha\rightarrow\neg\OBL\neg\alpha$ meaning that if $\alpha$ is 
obligatory, then its opposite ($\neg\alpha$) is not. Prohibitions can 
modelled as negative obligations, thus $\alpha$ is forbidden if its 
opposite is obligatory, that is $\OBL\neg\alpha$. Furthermore, it has 
been argued that maintenance  obligations are suitable to model prohibitions.  

Based on the discussion above, considering that the normative constraints
in the scenario of Section \ref{sec:scenario} are actually prohibitions, 
we formalise the scenario using $\G$ for maintenance obligations (actually 
prohibitions) and $\F$ for permissions.  We temporarily suspend judgement  
whether using an operator suitable to model achievement obligations to 
model the  dual permission for maintenance obligation is appropriate or not. 
All we remark here is that any formalism meant to model real life norms should 
account for both obligations and permissions as first class citizens.
 
A first possible \emph{prima facie} formalisation of the conditions set out 
in the Privacy Act is:
\begin{enumerate}
\item $\G\neg A$, $(\G\neg A\wedge A)\rightarrow \G B$;
\item $C \rightarrow \F A$;
\item $\G\neg D$;
\item $\F A\rightarrow \F D$.
\end{enumerate}
The set of formulas above exhibits some problems. First of all, in a situation 
where we have $C$ we get a contradiction from 1. and 2., i.e., $\G\neg A$
and $\F A$, and then a second from 3., and  2. and 4., namely $\G\neg D$ and 
$\F D$. This is due to the fact that normative reasoning is defeasible. 
Shortly and roughly a conclusion can be asserted unless there are reasons against 
it.  In addition, to get the expected results, we have to consider that the scenario
uses strong permissions, where the permissions derogates the obligations to the 
contrary, or, in other terms, that the permissions are exceptions to the 
obligations. To accomplish this we have to specify that 2. \emph{overrides} 1., 
and 4. \emph{overrides} 3. Technically, the overrides relationship can be 
achieved using the following procedure:\footnote{The focus of this paper is not 
how to implement defeasibility or non-monotonicity in LTL or in another monotonic
logic, thus we just exemplify a possible procedure.}
\begin{enumerate}
\item rewrite the formulas involved as conditionals. Thus $\G\neg A$ can be 
  rewritten as $\top \rightarrow \G\neg A$.
\item add the negation of the antecedent of the overriding formulas to the 
  antecedent of the formulas overridden formula. Accordingly 
  $\top \rightarrow \G\neg A$ is transformed into 
  $\neg C\rightarrow \G\neg A$.\footnote{A side-effect of this procedure, 
    which is harmless for the purpose of this paper, is that now the 
    combination of 3. and 4. makes $\F A$ and $\F D$ equivalent, namely 
    $\F A \equiv\F D$.}
\end{enumerate}
The second aspect we concentrate on is the form of the formulas in 1., 
in particular on the expression 
\begin{equation}\label{eq:ltl-violation}
(\G\neg A\land A) \rightarrow \G B.
\end{equation}
To start with they bear resemblance with the so called \emph{contrary-to-duty
obligations}. A contrary-to-duty obligation states that an 
obligation/prohibition is in force when the opposite of an obligation/prohibition
holds. The template for contrary-to-duty obligations is given by the pair
(a) $\OBL\alpha$ and (b) $\neg\alpha\rightarrow\OBL\beta$. Contrary-to-duty 
obligations are typically problematic for deontic logic and the source of 
inspiration for a wealth of research in the field (see 
\cite{prakkensergot96,carmo2002deontic}). The formula under scrutiny is indeed
related, but there is a difference: it explicitly requires a violation, while
the structure in (b) does not. In the context of the Privacy Act scenario (b)
would mean that an entity has the obligation to destroy collected personal 
information without accessing simply because they collected it (even in 
the case the collection was legal, or even when they had the mandate to 
collect it and eventually preserve it).

Accordingly, we introduce the class of \emph{compensatory} (contrary-to-duty)
\emph{obligations}. A compensatory obligation states that an 
obligation/prohibition is in force as the result of the violation of another 
obligation/prohibition. Thus the obligation triggered in 
response to the violation (secondary obligation) compensates the violation of the 
violated obligation (primary obligation). In other words a situation where the primary
obligation is violated, but the secondary obligation is fulfilled is still 
deemed legal, even if it is less ideal than the case where the primary 
obligation is fulfilled.\footnote{We do not exclude the case that there are 
situations where norms have the form of what we call compensatory obligations, 
but where the obligation in response to the violation does not (legally) 
compensate the  violation.} The language employed in the Privacy Act suggests 
that that the conditions stated in Section 1 and Section 2 of the Act 
correspond to a case of compensatory obligation. 

We turn now our attention to the issue of how to formalise compensatory 
obligations in LTL.  The  first concern we have when we look at 
\ref{eq:ltl-violation} we notice that its antecedent is always false, i.e., 
$\G\neg A\wedge A\equiv\bot$, since $\G\neg A$ implies that $A$ is false 
in all worlds following the world where the formula is evaluated including 
that world, but at the same time $A$ is required to be true at that world. 
The second issue is that the compensation is assumed to be a maintenance
obligation while the textual provision suggests it is a achievement 
obligation. We shortly discuss that achievement obligation should be 
represented by $\F$, but $\F$ is used to model permissions. 

To avoid the issues just discussed we introduce a new binary (temporal) 
operator $\otimes$ for compensatory obligations\footnote{The idea of using a 
specific operator for compensatory (contrary-to-duty) obligations is presented 
in \cite{ajl:ctd}.}. What we have to do for this end is to identify
the conditions under which a maintenance obligation is violated. The 
maintenance obligation $\OBL\alpha$ is violated if there is a instant in the
interval of validity of the obligation where $\alpha$ does not hold, namely
$\neg\alpha$ holds. The second thing is to define what it means to compensate
a violation. Suppose that we are told that the violation of $\alpha$ is 
compensated by $\beta$. A natural intuition for this is that there is an instant 
in the interval of validity of $\OBL\alpha$ where $\neg\alpha$ holds, and there 
is an instant successive to the violation where the course of action described 
by $\beta$ holds. Based on the intuition just described LTL seems well suited to this task. Here is the evaluation condition for 
$\otimes$:\footnote{Again the focus of the paper is not on how to 
properly model compensatory (contrary-to-duty) obligations. The operator 
presented here does its job in the context of the paper. For alternative 
definitions in the context of temporal logic or inspired by temporal logic 
see \cite{piolle:10,Grossi:2013}. For a semantic approach not based on 
temporal logic see \cite{deon2014semantics}.}\textsuperscript{,}\footnote{This condition 
implements compensatory obligations when the primary obligation is a 
maintenance obligation and the secondary obligation is an achievement 
obligation. Similar definitions can be given for other combinations of 
primary and secondary obligations.}
\begin{equation}
TS,\sigma \models \phi\otimes\psi \mbox{ iff }
  \forall i\geq 0,\ TS,\sigma_i\models \phi; \mbox{ or } 
  \exists j,k: 0\leq j\leq k, \ TS,\sigma_j\models \neg\phi \mbox{ and }  
  TS,\sigma_k\models \psi.
\end{equation}
We are now ready to provide the formalisation of the Privacy Act.   
\begin{enumerate}[label={N\arabic*.}]
\item $\neg C \rightarrow (\neg A\otimes B)$;
\item $C \rightarrow \F A$;
\item $\G \neg A\rightarrow\G \neg D$;
\item $\F A \rightarrow \F D$.
\end{enumerate}
Transition systems can be use to model runs of systems, possible ways in
which business processes can be executed, the actions of an agents or 
more in general the dynamic evolution of a system or the world. Norms are 
meant to regulate the behaviour of systems, how organisations run their
business, the actions of agents and so on.  So, how do we check if a 
particular course of actions (modelled by a transition system) 
complies with a set of norms (where the norms are formalised in LTL)?
Simply, if the transition system is a model for the set of formulas
representing the norms.
 
Consider a transition system $TS= \langle S,R,v\rangle$ where
\begin{enumerate}
  \item $S=\set{t_{i}\colon i\in\Nat}$,
  \item $R=\set{(t_i,t_{i+1})\colon i\in\Nat}$,
  \item $\neg C\in v(t_i)$ for all $i\in\Nat$, $A\in v(t_1)$, $D\in v(t_1)$ 
    and $B\in v(t_2)$.
\end{enumerate}
The transition system is such that 
\begin{equation}
 TS,t_i\models \neg C, \qquad 
 TS,t_1\models A, \qquad 
 TS,t_1\models D, \qquad
 TS,t_2\models B.
\end{equation}
This transition system implements the scenario where at no time there is a 
Court Order authorising the collection of personal information ($\neg C$ 
for all $t_i$), an entity collects personal information ($A$ at time $t_1$) 
and successively destroys it ($B$ at time $t_2$), and at the same time when 
personal information was collected medical information was collected ($D$ 
at time $t_1$).

It is immediate to verify that the transition system $TS$ is a model of 
N1--N4, namely:
\begin{equation}
  \forall t\in S\colon TS,t\models 
    \text{N1}\wedge\text{N2}\wedge\text{N3}\wedge\text{N4}.
\end{equation}
Accordingly, $TS$ is compliant with N1--N4. However, there is state $t_1$ 
where both $\neg C$ and $D$ hold. In Section \ref{sec:scenario} we argued 
that a situation where $\neg C$ and $D$ both hold is not compliant. 
Therefore, we have a paradox, the formalisation indicates that the scenario
is compliant, the course of actions described by the transition system does 
not result in a contradiction, so no illegal action is performed (or better, 
the collection of personal information is illegal, but its compensation, 
destruction of the personal information, makes full amends to it), but our legal 
intuition suggests that the collection of medical information in the 
circumstances of the scenario is illegal.\footnote{We run a pseudo empirical
validation of the scenario by proposing the scenario and the Privacy Act to about 
a dozen legal professionals ranging from corporate legal councillors, 
to high court judges to law professors. They all agree without any hesitation 
that the collection of medical information under the circumstances described 
by the scenario is illegal. However, a true validation can be only given 
either by a law court adjudication of a case where the norms at hand are 
isomorphic to the Privacy Act, or by any body with the power to give a
true interpretation of an act isomorphic to the act we proposed for the 
scenario.}

\section{Conclusion}
\label{sec:conclusion}
The contribution of this note is twofold. First we presented a novel paradox
for Deontic Logic inspired by real life norms. In particular the logical 
structures used in the paradox appear frequently in real life (legal) norms.
The second contribution was a short analysis of how to represent norms in
Linear Temporal Logic, and that the proposed formalisation results in a 
paradox, showing that LTL might not be suitable to model norms and legal
reasoning.

We would like to point out that the discussion in the previous section just 
shows that \emph{a} particular formalisation based on LTL is not suitable 
to represent the scenario, not that LTL per se is not able to represent the 
scenario. Indeed one could create all possible full paths in a transition 
system not breaching the norms, and then using the paths to synthesise the 
norms that regulate the transition system. However, we believe that such 
\emph{ex post} analysis is useless. First humans have to perform the 
reasoning to determine which norms hold and when and then which paths violate 
the norms. In addition the strength of LTL is the ability to verify 
specifications against transition systems. But in such a case, given that
the specifications are derived from the transition systems, the verification
is always positive and totally uninformative. Furthermore, we believe that
the formalisation we proposed, while naive, is extremely intuitive. The 
major objection, as we remarked in Section \ref{sec:scenario}, is that
permissions are modelled using $\F$, and we hinted that $\F$ might be
suitable to model achievement obligation, and using a particular type 
of obligation to model permissions is not appropriate and counter-intuitive 
outcomes are to be expected. We fully agree with this objection, but if we
agree that a permission is the lack of an obligation to the contrary, then $\F$ 
is the natural choice for permissions for prohibition (maintenance obligations). 
The other issue is that if we do not use $\F$, the issue is how to model
permission, and the alternative is that LTL does not support permissions. 
The act we presented clearly shows that there are acts where permissions must
be represented and that permissions play an important role in determining
which obligations are in force and when they are in force. Hence, any 
formalisation excluding permission is doomed to be unable to represent the
vast majority of real life legal norms.  

   
The final remark we want to make is that the paradox is not restricted to 
LTL. It can be easily replicated in Standard Deontic Logic (and it is
well know that Standard Deontic Logic is plagued with many other 
contrary-to-duty paradoxes).  A root-cause analysis of the paradox is that 
a violation of a compensable obligation results in a sub-ideal state. Hence, 
there is a state with a violation that is still deemed legal. This means, 
that there is a (somehow) legal state, and if permission is evaluated as being 
in at least one legal state, then the violation has to be evaluated as (somehow) 
permitted. Part of the problem is that in such somehow legal states there 
might be other true legitimate permissions which are not the violation of 
compensable obligations.  Accordingly, we conjecture, that logics using truth
of a formula in at least one (somehow) legal state to determine whether 
something is permitted have counterparts of the paradox we presented. However, 
a careful analysis of existing deontic logics is needed to evaluate if they
are actually affected by the paradox.

\subsection*{Acknowledgements}


I thank Antonino Rotolo and Giovanni Sartor for fruitful comments on 
previous drafts of this paper. I also thank the participants to NorMAS 
2014 for the discussions and valuable suggestions.

NICTA is funded by the Australian Government through the Department of 
Communications and the Australian Research Council through the ICT Centre 
of Excellence Program.


\printbibliography

\end{document}